\documentclass[runningheads]{llncs}
\usepackage{graphicx}
\usepackage{subcaption}
\usepackage{xcolor}
\usepackage{cite}
\usepackage{xspace}
\usepackage{amsmath}
\usepackage{amsfonts}
\usepackage[belowskip=-15pt,aboveskip=0pt]{caption}

\def\UncertainDeepSSM{Uncertain-DeepSSM\@\xspace}

\newcommand{\etal}{\textit{et al.}~}
\newcommand{\ie}{i.e.,~}

\newcommand{\eg}{e.g.,~}
\newcommand{\aka}{a.k.a.~}

\newcommand{\x}{\mathbf{x}}
\newcommand{\y}{\mathbf{y}}
\newcommand{\z}{\mathbf{z}}
\renewcommand{\a}{\mathbf{a}}
\newcommand{\e}{\mathbf{e}}

\newcommand{\zb}{\bar{\mathbf{z}}}

\newcommand{\ati}{\widetilde{\mathbf{a}}_i}
\newcommand{\atil}{\widetilde{a}_{il}}

\newcommand{\xn}{\x_n}
\newcommand{\yn}{\y_n}
\newcommand{\zn}{\z_n}
\newcommand{\an}{\a_n}
\newcommand{\en}{\e_n}

\newcommand{\xs}{\x_s}
\newcommand{\ys}{\y_s}
\newcommand{\zs}{\z_s}

\renewcommand{\xi}{\x_i}
\newcommand{\yi}{\y_i}
\newcommand{\zi}{\z_i}
\newcommand{\zbi}{\zb_i}
\newcommand{\ai}{\a_i}

\newcommand{\ail}{a_{il}}
\newcommand{\eil}{e_{il}}
\newcommand{\zbil}{\bar{z}_{il}}

\newcommand{\R}[1]{\mathbb{R}^{#1}}
\newcommand{\Ru}[1]{\mathbb{R}_{#1}}

\newcommand{\U}{\mathbf{U}}

\newcommand{\bmu}{\boldsymbol{\mu}}
\newcommand{\bDelta}{\boldsymbol{\Delta}}
\newcommand{\bTheta}{\boldsymbol{\Theta}}

\newcommand{\Dset}{\mathcal{D}}
\renewcommand{\L}{\mathcal{L}}

\newcommand{\diag}{\operatorname{diag}}

\begin{document}
\title{\UncertainDeepSSM: From Images to Probabilistic Shape Models}
\titlerunning{\UncertainDeepSSM}
\author{Jadie Adams\inst{1,2} \and
Riddhish Bhalodia\inst{1,2} \and
Shireen Elhabian\inst{1,2}}
\authorrunning{J. Adams et al.}
\institute{Scientific Computing and Imaging Institute, University of Utah, UT, USA \and
School of Computing, University of Utah, UT, USA}
\maketitle
\begin{abstract}

Statistical shape modeling (SSM) has recently taken advantage of advances in deep learning to alleviate the need for a time-consuming and expert-driven workflow of anatomy segmentation, shape registration, and the optimization of population-level shape representations. DeepSSM is an end-to-end deep learning approach that extracts statistical shape representation directly from unsegmented images with little manual overhead. It performs comparably with state-of-the-art shape modeling methods for estimating morphologies that are viable for subsequent downstream tasks. Nonetheless, DeepSSM produces an overconfident estimate of shape that cannot be blindly assumed to be accurate. Hence, conveying what DeepSSM does not know, via quantifying granular estimates of uncertainty, is critical for its direct clinical application as an on-demand diagnostic tool to determine how trustworthy the model output is. Here, we propose \UncertainDeepSSM as a unified model that quantifies both, data-dependent aleatoric uncertainty by adapting the network to predict intrinsic input variance, and model-dependent epistemic uncertainty via a Monte Carlo dropout sampling to approximate a variational distribution over the network parameters. Experiments show an accuracy improvement over DeepSSM while maintaining the same benefits of being end-to-end with little pre-processing. 

\keywords{Uncertainty Quantification \and Statistical Shape Modeling \and Bayesian Deep Learning}
\end{abstract}
\section{Introduction} 
Morphometrics and its new generation, statistical shape modeling (SSM), have evolved into an indispensable tool in medical and biological sciences to study anatomical forms. SSM has enabled a wide range of biomedical and clinical applications (\eg \cite{greig2001brain,harris2013cam,cates2013afib,bhalodia2018endtoend,bhalodia2020quantifying,cates2014computational,kozic2010optimisation,galloway2013large,bryan2009use,zhao2008hippocampus,wang2012comprehensive}).
Morphological analysis requires parsing the anatomy into a quantitative representation consistent across the population at hand to facilitate the testing of biologically relevant hypotheses. 
A popular choice for such a representation is using \textit{landmarks} that are defined consistently using invariant points, \ie \textit{correspondences}, across populations \cite{sarkalkan2014statistical}. 
Coordinate transformations (\eg \cite{RTW:Jos04,RTW:Jos2000}) hold promise as an alternative representation, but the challenge is finding the anatomically-relevant transformation that quantifies differences among shapes.
Ideally, landmarking is performed by anatomy experts to mark distinct, and typically few anatomical features \cite{baccetti1999thin,lamecker2002statistical}, but it is time-intensive and cost-prohibitive, especially for 3D images and large cohorts. More recently, dense sets of correspondence points that capture population statistics are used, thanks to advances in computationally driven approaches for shape modeling (\eg  \cite{cates2017shapeworks,cates2007shape,davies2002MDL,styner2006spharm,durrleman2014morphometry}).

Traditional computational approaches to automatically generate dense correspondence models, \aka point distribution models (PDMs), still entail a time-consuming, expert-driven, and error-prone workflow of segmenting anatomies from volumetric images, followed by a processing pipeline of shape registration, correspondence optimization, and projecting points onto some low-dimensional shape space for subsequent statistical analysis. Many of these steps require significant parameter tuning and/or quality control by the users.
The excessive time and effort to construct population-specific shape models have motivated the use of deep networks and their inherent ability to learn complex functional mappings to regress shape information directly from images and incorporate prior knowledge of shapes in image segmentation tasks (\eg \cite{DeepSSM,huang2017heartnet,milletari2017stats,xie2017deepshape,zheng2015detection}). 
However, deep learning in this context has drawbacks. Training deep networks on volumetric images is often confounded by the combination of high-dimensional image spaces and limited availability of training images labeled with shape information. Additionally, deep networks can make poor predictions with no indication of uncertainty when the training data weakly represents the input. 
Computationally efficient automated morphology assessment when integrated with new clinical tools as well as surgical procedure, has potential to improve medical care standards and clinical decision making. However, uncertainty quantification is a must in such scenarios, as it will allow professionals to determine the trustworthiness of such a tool and prevent unsafe predictions.
\begin{figure}[!ht]
    \centering
    \includegraphics[width=\textwidth]{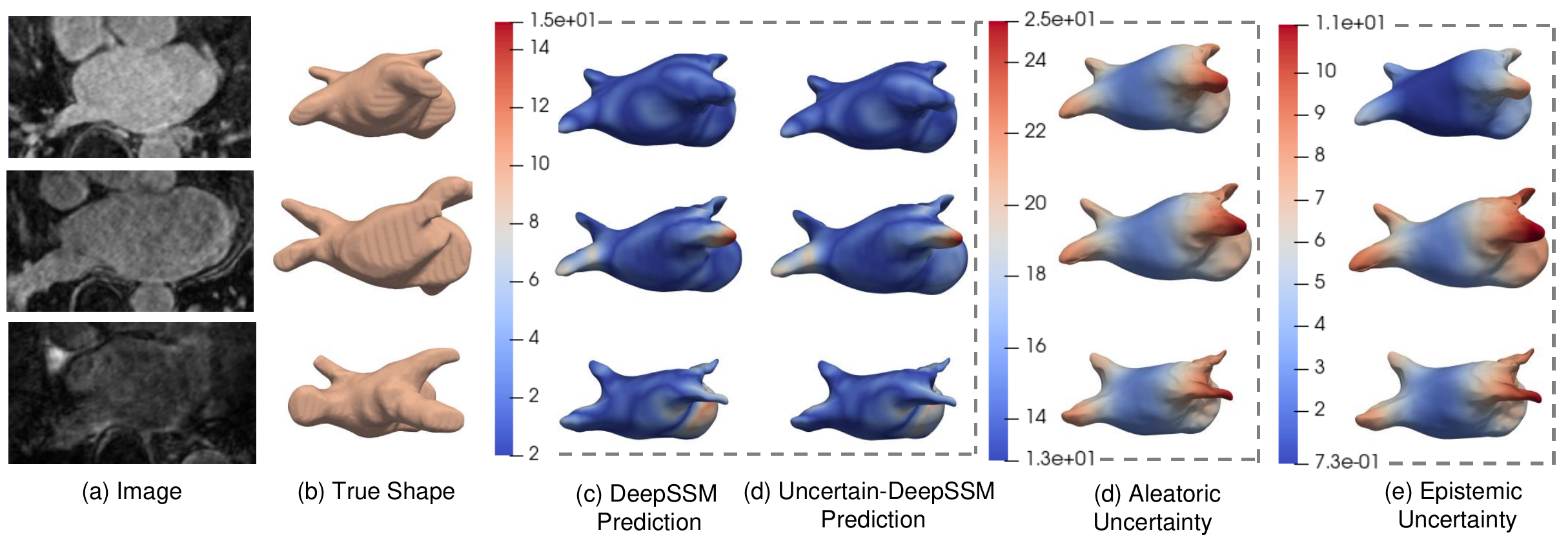}
    \caption{\textbf{Shape prediction and uncertainty quantification} on left atrium MRI scans. The images (a) are input and the true shapes (surface meshes) (b) are from ground truth segmentations. 
    Shapes in (c) and (d) are constructed from DeepSSM and  \UncertainDeepSSM predictions, respectively.
    The heat maps on surface meshes in (c) and (d) show the surface-to-surface distance to (b) (the error in mm). The aleatoric (d) and epistemic (e) output from \UncertainDeepSSM are shown as heat maps on the predicted mesh. Our model outputs increased uncertainty where error is high.}
    \label{fig:teaser}
\end{figure}

Here, we focus on a particular instance of a deep learning-based framework, namely DeepSSM \cite{DeepSSM}, that maps unsegmented 3D images to a low-dimensional shape descriptor. 
Mapping to a low-dimensional manifold, compared with regressing correspondence points, has a regularization effect that compensates for misleading image information and provides a topology-preserving prior to shape estimation.
DeepSSM also entails a population-driven data augmentation approach that addresses limited training data, which is typical with small and large-scale shape variability.
DeepSSM has been proven effective in characterizing pathology  \cite{DeepSSM} and performs statistically similar to traditional SSM methods in downstream tasks such as disease recurrence predictions \cite{bhalodia2018endtoend}.
Nonetheless, DeepSSM, like other deep learning-based frameworks predicting shape, produces an overconfident estimate of shape that can not be blindly assumed to be accurate. Furthermore, the statistic-preserving data augmentation is bounded by what the finite set of training samples captures about the underlying data distribution.
In this paper, we formalize \UncertainDeepSSM, a unified solution to limited training images with dense correspondences and model prediction over-confidence.
\UncertainDeepSSM quantifies granular estimates of uncertainty with respect to a low-dimensional shape descriptor to provide spatially-coherent, localized uncertainty measures (see Fig. \ref{fig:teaser}) that are robust to misconstruing factors that would typically affect point-wise regression, such as heterogeneous image intensities and noisy diffuse shape boundaries.
\UncertainDeepSSM produces probabilistic shape models directly from 3D images, conveying what DeepSSM does not know about the input and providing an accuracy improvement over DeepSSM while being end-to-end with little required pre-processing. 
\section{Related Work}

DeepSSM \cite{DeepSSM} is based on works that show the efficacy of convolutional neural networks (CNNs) to extract shape information from images.
Huang \etal \cite{huang2017heartnet} regress shape orientation and position conditioned on 2D ultrasound images. Milletari \etal \cite{milletari2017stats} segment ultrasound images using low-dimensional shape representation in the form of principal component analysis (PCA) to regress landmark positions. Oktay \etal \cite{ACNN} incorporate prior knowledge about organ shape and location into a deep network to anatomically constrain resulting segmentations. However, these works provide a point-estimate solution for the task at hand.

In Bayesian modeling, there are two types of uncertainties \cite{Kiureghian2009AleatoryOE,Kendall2017}.
\textit{Aleatoric} (or data) uncertainty captures the uncertainty inherent in the input data, such as over-exposure, noise, and the lack of the image-based features indicative of shapes.
\textit{Epistemic} (or model) uncertainty accounts for uncertainty in the model parameters and can be explained away, given enough training data \cite{Kendall2017}. 

Aleatoric uncertainty can be captured by placing a distribution over the model output. In image segmentation tasks, this has been achieved by sampling segmentations from an estimated posterior distribution \cite{chang2011efficient,le2016sampling} and using conditional normalizing flows \cite{selvan2020uncertainty} to infer a distribution of plausible segmentations conditioned on the input image. These efforts succeed in providing shape segmentation with aleatoric uncertainty measures, but do not provide a shape representation that can be readily used for population-level statistical analyses.
T{\'{o}}thov{\'{a}} \etal \cite{Parisot2018} incorporate prior shape information into a deep network in the form of a PCA model to reconstruct surfaces from 2D images with an aleatoric uncertainty measure that is quantified via conditional probability estimation. Besides being limited to 2D images, quantifying point-wise aleatoric uncertainty makes this measure prone to inherent noise in images.

Epistemic uncertainty is more difficult to model as it requires placing distributions over models rather than their output. Bayesian neural networks \cite{mackay1992practical,denker1991transforming,gal2016uncertainty} achieve this by placing a prior over the model parameters, then quantifying their variability. Monte Carlo dropout sampling, which places a Bernoulli distribution over model parameters \cite{gal2015bayesian}, has effectively been formalized as a Bayesian approximation for capturing epistemic uncertainty \cite{gal2016dropout}. Aleatoric and epistemic uncertainty measures have been combined in one model for tasks such as semantic segmentation, depth regression, classification, and image translation \cite{Kendall2017,kwon2018uncertainty,reinhold2020validating}, but never for SSM.

\UncertainDeepSSM produces probabilistic shape models directly from images that quantifies both the data-dependent aleatoric uncertainty and the model-dependent epistemic uncertainty.
We quantify aleatoric uncertainty by adapting the network to predict intrinsic input variance in the form of mean and variance for the PCA scores and updating the loss function accordingly \cite{nix1994estimating,le2005heteroscedastic}. This enables explicit modeling of the heteroscedastic-type of aleatoric uncertainty, which is dependent on the input data.
We model epistemic uncertainty via a Monte Carlo dropout sampling to approximate a variational distribution over the network parameters by sampling PCA score predictions with various dropout masks. 
This approach provides both uncertainty measures for each PCA score that are then mapped back to the shape space for interpretable visualization. Uncertainty fields on estimated 3D shapes convey insights for how the given input relate to what the model knows. For instance, such uncertainties could help pre-screen for pathology if \UncertainDeepSSM is trained on controls. 
Furthermore, explicit modeling of uncertainties in \UncertainDeepSSM provides more accurate predictions, compared with DeepSSM, with no additional training steps. This indicates the ability of \UncertainDeepSSM to better generalize in limited training data setting.

\section{Methods}

A trained \UncertainDeepSSM model provides shape descriptors, specifically PCA scores, with uncertainty measures directly from 3D images (\eg CT, MRI) of anatomies.
In this section, we describe the data augmentation method, the network architecture, training strategy, and uncertainty quantification.

\subsection{Notations}

Consider a paired dataset $\{ (\xn,\yn) \}_{n=1}^{N}$ of $N$ 3D images $\yn \in \R{H \times W \times D}$ and their corresponding shapes $\xn \in \R{3M}$, where each shape is represented by $M$ 3D correspondence points.
We generate a PDM from segmentations. This entails the typical SSM pipeline that includes pre-processing steps (registration, resampling, smoothing, ...), and correspondence (\ie PDM) optimization. 
In practice, any PDM generation algorithm can be employed. Here, we use the open-source \textit{ShapeWorks} software \cite{cates2017shapeworks} to optimize surface correspondences using anatomies segmented from the \textit{training} images. 
Next, high-dimensional shapes (\ie PDM) in the shape space (of dimension $\mathbb{R}^{3M}$) are mapped to low-dimensional PCA scores $\z \in \R{L}$ in the PCA subspace that is parameterized by a mean vector $\bmu \in \R{3M}$, a diagonal matrix of eigen values $\bDelta \in \R{L \times L}$, and a matrix of eigen vectors $\U \in \R{3M \times L}$, where $\z = \U^T(\x -\bmu)$ and $L \ll 3M$ is chosen such that at least 95\% of the population variation is explained. 
The PCA scores $\zn$ associated with the training image $\yn$ serve as a supervised target to be inferred by the \UncertainDeepSSM~network and mapped deterministically to correspondence points, where $\xn = \U\zn + \bmu$.
The network thus defines a functional map $f_{\bTheta}: \R{H \times W \times D} \rightarrow \R{L}$ that is parameterized by the network parameters $\bTheta$, where $\z = f_{\bTheta}(\y)$.
Uncertainties are quantified in the PCA subspace, such that the PCA scores of the $n-$th training shape $\zn$ is associated with vectors of aleatoric variances $\an \in \R{L}_{+}$ and epistemic variances $\en \in \R{L}_{+}$.
\begin{figure}
    \begin{center}
    \includegraphics[width=\textwidth]{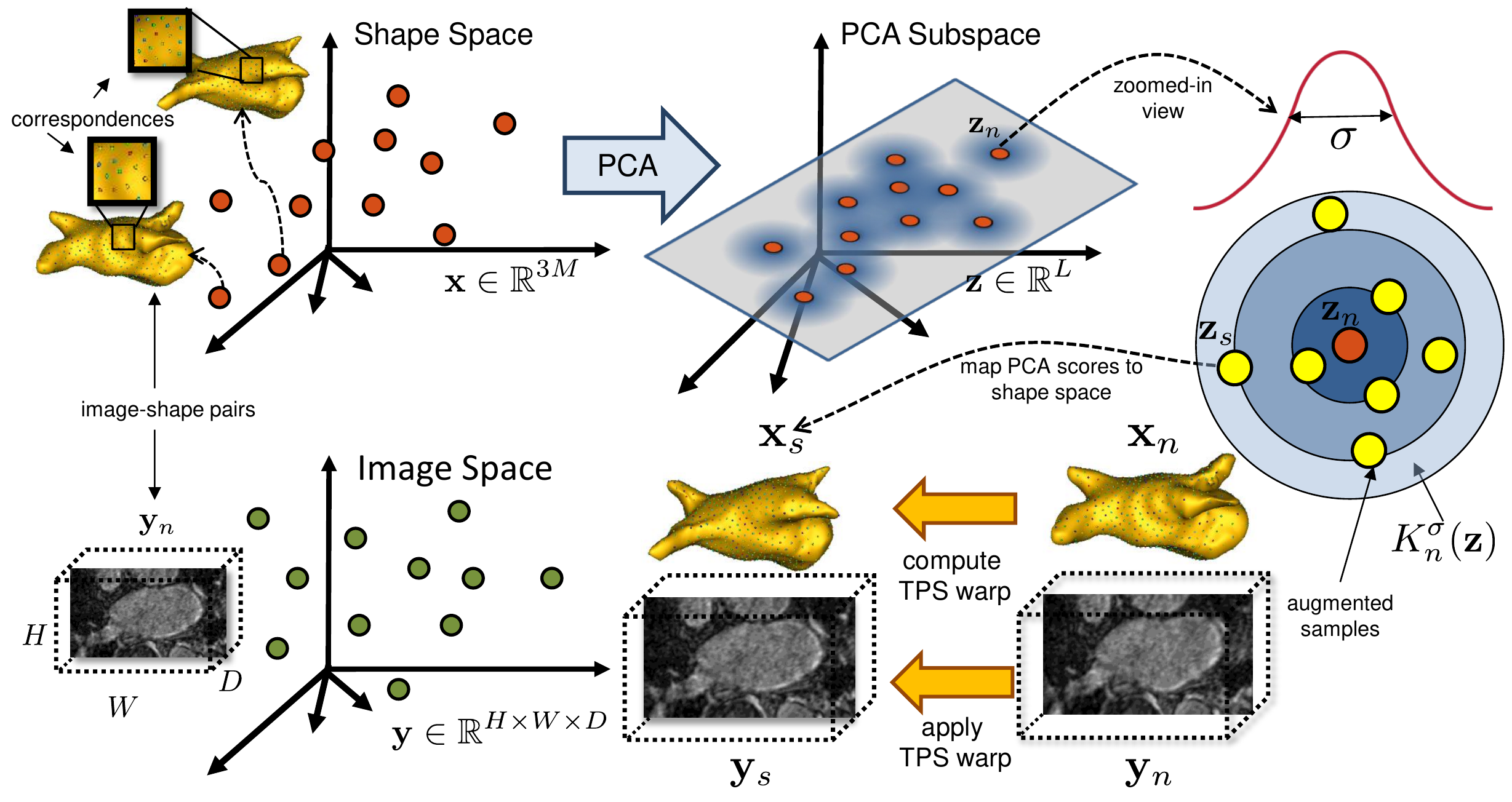}
    \end{center}
    \caption{\textbf{Data augmentation:} PCA is used to compute the PCA scores $\{\zn\}_{n=1}^N$ of the training shape samples $\{\xn\}_{n=1}^N$. Augmented samples $\zs$ are randomly drawn from a collection of multivariate Gaussian distributions $K_\sigma^n(\mathbf{z})$, each with a training example $\zn$ as the mean and covariance $\sigma^2\mathbf{I}_L$. The correspondences $\xs$ are used to compute a TPS warp to map $\xn$ to $\xs$, which is used to warp the respective image $\yn$ to a new image $\ys$ with known shape parameters $\zs$.
    }
    \label{fig:augmentation}
\end{figure} 

\subsection{Data Augmentation}\label{sect:data_aug}

DeepSSM augments training data with shape samples generated from a single multivariate Gaussian distribution in the PCA subspace, where an add-reject strategy is employed to prevent outliers from being sampled.
We use a kernel density estimate (KDE) instead, where augmented samples $\zs$ are drawn from $p_\sigma(\z)$ in \eqref{eqn:pz}. This probability density better captures the underlying shape distribution by not over-estimating the variance and avoiding sampling implausible shapes from high probability regions in case of multi-modal distributions. 
\begin{equation}\label{eqn:pz}
  p_\sigma(\z) =\frac{1}{N} \sum_{n=1}^N K_n^\sigma(\z), ~~\operatorname{s.t.}~~K_n^\sigma(\z) = \frac{1}{(2 \pi \sigma^2)^{L/2}}  \exp \left( - \frac{||\z - \zn||^2} {2\sigma^2}\right),
\end{equation}
where $\sigma \in \Ru{+}$ denotes the kernel bandwidth and is computed as the average nearest neighbor distance in the PCA subspace, \ie $\sigma^2 = \frac{1}{N} \sum_n \operatorname{min}_{k \neq n} (\zn - \z_k)^T \bDelta^{-1} (\zn - \z_k)$. 
As illustrated in Fig. \ref{fig:augmentation}, a sampled vector of PCA scores $\zs \in \R{L}$ from the kernel of the $n-$th training sample $K_n^\sigma(\z)$ is mapped to correspondence points $\xs \in \R{3M}$, where  $\xs = \U\zs+ \bmu$. 
Using the $\xn \leftrightarrow \xs$ correspondences, we compute thin-plate spline (TPS) warp \cite{bookstein1989principal} to deform the image $\yn$ and construct the augmented image $\ys$. 
With this augmentation method, we can construct an augmented training set $\{ (\xs,\ys) \}_{s=1}^{S}$ of $S$ 3D images, their corresponding shapes, and the supervised targets $\{ \zs \}_{s=1}^{S}$, which respects the population-level shape statistics and the intensity profiles of the original dataset.
 
\subsection{Adaptations for Uncertainty Quantification}

We extend the network architecture and loss function of DeepSSM to estimate both types of uncertainties and the shape descriptor in the form of PCA scores.

\vspace{0.05in}
\noindent\textbf{Heteroscedastic aleatoric uncertainty} is a measure of data uncertainty, and hence can be learned as a function of the input. Given a training set $\Dset = \{(\yi,\zi)\}_{i=1}^I$ that includes both real and augmented samples, where $I = N + S$, DeepSSM is trained to minimize the L2 loss between groundtruth $\zi$ and predicted $f_{\bTheta}(\yi)$.
In \UncertainDeepSSM, the network architecture is modified to estimate both the mean $\zbi$ and variance $\ai$ of the PCA scores, where $[\zbi, \ai] = f_{\bTheta}(\yi)$. The variance acts as an uncertainty regularization term that does not require a supervised target since it is learned implicitly through supervising the regression task. For training purposes, we let the network predict the log of the variance, $\atil = \log \ail^2$, where $\ail$ captures the aleatoric uncertainty along the $l-$th PCA mode of variation. This forces the variance to be positive and removes the potential for division by zero.  \UncertainDeepSSM~is thus trained to minimize the Bayesian loss in \eqref{eqn:bayesloss}, where $\zbi = f_{\bTheta}^{\z}(\yi)$ and $\ati = f_{\bTheta}^{\a}(\yi)$ are the $\z-$ and $\a-$ outputs of the network, respectively (see Fig. \ref{fig:architecture}). 
 \begin{equation}\label{eqn:bayesloss}
 \L(\bTheta) = \frac{1}{2LI} \sum_{i=1}^I \left\{ \left[\zi - \zbi\right]^T \diag\left(\exp(\ati)\right)^{-1} \left[\zi - \zbi\right] +  \sum_{l=1}^L \atil \right\}
 \end{equation}
The second term in \eqref{eqn:bayesloss} learns a loss attenuation, preventing the network from predicting infinite variance for all scores. 

\vspace{0.05in}
\noindent \textbf{Epistemic uncertainty} is a measure of the model's ignorance that can be quantified by modeling distributions over the model parameters $\bTheta$. We place a Bernoulli distribution over network weights by making use of the Monte Carlo dropout sampling technique \cite{gal2015bayesian, gal2016dropout}. In particular, a dropout layer with a probability $\kappa$ is added to every layer (convolutional and fully connected) in the \UncertainDeepSSM~network (Fig. \ref{fig:architecture}). Dropout is used in both, training and testing, where in testing, it is used to sample from the approximate posterior. The statistics of the distribution of network predictions with different dropout masks reflects the model’s epistemic uncertainty. Consider $V$ dropout samples, the epistemic uncertainty of the $l-$th PCA mode of variation is computed as,
\begin{equation}\label{eqn:epistemic}
\eil = \frac{1}{V} \sum_{v=1}^V \left(\zbil^{(v)}\right)^2 - \left(\frac{1}{V} \sum_{v=1}^V \zbil^{(v)} \right)^2
\end{equation}
\noindent where $\zbi^{(v)} = f_{\bTheta_v \sim p(\bTheta)}^{\z}(\yi)$ is the $\z-$output of the network for the randomly masked network parameters $\bTheta_v$.

\subsection{Architecture and Training}

\begin{figure}
    \centering
    \includegraphics[width=\textwidth]{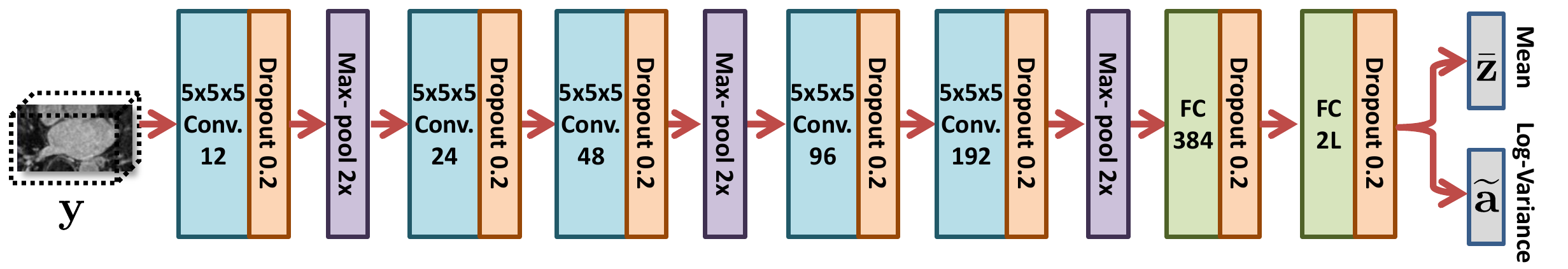}
    \caption{\UncertainDeepSSM network architecture}
    \label{fig:architecture}
\end{figure}

The network architecture of \UncertainDeepSSM (Fig. \ref{fig:architecture}) is similar to DeepSSM with five convolution layers followed by two fully connected layers. However in \UncertainDeepSSM, dropout is added and batch normalization is removed. Combining batch normalization and dropout leads to a variance shift that causes training instability \cite{disharmony}. Hence, we normalize the input images to compensate for not using batch normalization.
The PCA scores are also whitened to prevent the model from favoring the dominant PCA modes. A dropout layer with a probability of $\kappa = 0.2$ is added after every convolutional and fully connected layer.
Data augmentation is used to create a set of $I = 4000$ training images and $1000$ validation images for training the network. PyTorch is used in constructing and training DeepSSM with Adam optimization \cite{adam} and a learning rate of 0.0001. Parametric ReLU \cite{he2015rectifiers} activation is used and network parameters are initialized by Xavier initialization \cite{xavier2010initialization}. To train \UncertainDeepSSM, the L2 loss function is used for the first epoch and the Bayesian loss function \eqref{eqn:bayesloss} is used for all following epochs. This allows the network to learn based on the task alone before learning to quantify uncertainty, resulting in better predictions and a more stable training. 

\subsection{Testing and Uncertainty Analysis}

When testing, dropout remains on and predictions are sampled multiple times. The predicted PCA scores and aleatoric uncertainty measure are first un-whitened. Using the dropout samples, the epistemic uncertainty measure is computed using \eqref{eqn:epistemic} based on the un-whitened predicted scores. 
To compute the accuracy of the predictions, we first map PCA scores to correspondence points and compare the surface reconstructed from these points to the surface constructed from the ground truth segmentation.
For surface reconstruction, we use the point correspondences between the population mean and the correspondence points of the predicted PCA scores to define a TPS warp that deform the mean mesh (from ShapeWorks \cite{cates2017shapeworks}) to obtain the surface mesh for the predicted scores.
The error is then calculated as the average of the surface-to-surface distance from the predicted to ground truth mesh and that of ground truth to predicted mesh.

To visualize uncertainty measures on the predicted mesh, the location of each correspondence point is modeled as a Gaussian distribution. To fit these distributions, we sample PCA scores from a Gaussian with the predicted mean and desired variance (aleatoric or epistemic), then map them to the PDM space.
This provides us with a distribution over each correspondence point, with mean and entropy indicating the coordinates of the point and the associated uncertainty scalar, respectively. Interpolation is then used to interpolate uncertainty scalars defined on correspondence points to the full reconstructed mesh.
\section{Results}

We compare the shape predictions from DeepSSM and \UncertainDeepSSM on two 3D datasets; a toy dataset of parametric shapes (supershapes) as a proof-of-concept and a real world dataset of left atrium MRI scans. In both experiments, we create three different test sets: \textit{control}, \textit{aleatoric}, and \textit{epistemic}. The control test set is well represented under the training population,  whereas the epistemic and aleatoric test sets are not.
Examples with images that differ from the training images are chosen for the aleatoric set, as this suggests data uncertainty. The epistemic set is chosen to demonstrate model uncertainty by selecting examples with shapes that differ from those in the training set. 
The test sets are held out from the entire data augmentation and training process. It is important that test sets are not used to build the PDMs, such that they are not reflected in the population statistics captured in the PCA scores. Hence, we use surface-to-surface distances between meshes to quantify shape-based prediction errors since testing samples do not have optimized (ground truth) correspondences.

\subsection{Supershapes Dataset}

As a proof-of-concept, we construct a set of 3D supershapes shapes \cite{Gielis2013}, which are a family of parameterized shapes. A supershape is parameterized by three variables, one which determines the number of lobes in the shape (or the shapes group), and two which determine the curvature of the shape. 
To create the training and validation sets, we generate 5000 3-lobe shapes with randomly drawn curvature values (using a $\chi^2$ distribution). For each shape, a corresponding image of size $98 \times 98 \times 98$ is formed, where the intensities of the foreground and background are modeled as Gaussian distributions with different means but same variance. Additive Gaussian noise is added and the images are blurred with a Gaussian filter to mimic diffuse shape boundaries.  
In this case, \UncertainDeepSSM predicts a single PCA score, where the first dominant PCA mode captures ~99\% of the shape variability. 
\begin{figure}[!ht]
    \centering
    \includegraphics[width=\textwidth]{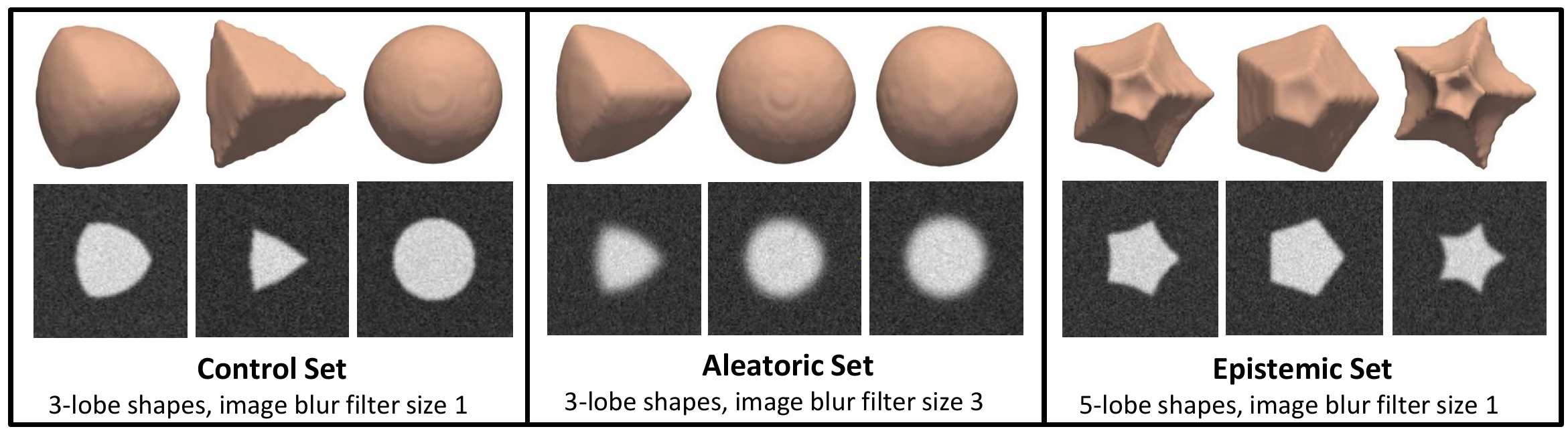}
    \caption{Examples of surfaces and corresponding image slices from supershapes test sets.}
    \label{fig:SS}
\end{figure}

We analyze output uncertainty measures on three different test sets, each of size 100. Examples of these can be seen in Fig. \ref{fig:SS}. The control test set is generated in the same manner as the training data and provides baseline uncertainty measures. The aleatoric test set contains shapes of the same shape group as the training, but the corresponding images are blurred with a larger Gaussian filter. This makes the shape boundary less clear, which has the effect of adding data uncertainty. For the epistemic test set, the images are blurred to the same degree as the training set, but the shapes belong to a different shape group. Here, we use 5-lobe shapes instead of 3-lobe to demonstrate model uncertainty. 
\begin{table}[h!]
    \begin{center}
    \resizebox{\textwidth}{!}{%
        \begin{tabular}{c|c|c|c|c|}
        \cline{2-5}
            & \multicolumn{1}{c|}{ \textbf{DeepSSM} }  & \multicolumn{3}{c|}{\textbf{\UncertainDeepSSM}} \\ \cline{2-5} 
            & \multicolumn{1}{c|}{\begin{tabular}[c]{@{}c@{}}\textbf{Surface-to-Surface}\\ \textbf{Distance}\end{tabular}} 
            & \multicolumn{1}{c|}{\begin{tabular}[c]{@{}c@{}}\textbf{Surface-to-Surface}\\ \textbf{Distance}\end{tabular}} 
            & \multicolumn{1}{c|}{\begin{tabular}[c]{@{}c@{}}\textbf{Aleatoric} \\ \textbf{Uncertainty}\end{tabular}} 
            & \multicolumn{1}{c|}{\begin{tabular}[c]{@{}c@{}}\textbf{Epistemic} \\ \textbf{Uncertainty}\end{tabular}} \\ \hline
        \multicolumn{1}{|r|}{\textbf{Control Test Set}}   & 0.670 $\pm$ 0.104 & 0.615 $\pm$ 0.163 & 7.413 $\pm$ 2.189 & 15.000 $\pm$ 11.762\\ \hline
        \multicolumn{1}{|r|}{\textbf{Aleatoric Test Set}} & 1.293 $\pm$ 0.679 & 0.798 $\pm$ 0.447 & 10.205 $\pm$ 2.276 & 22.178 $\pm$ 13.065 \\ \hline
        \multicolumn{1}{|r|}{\textbf{Epistemic Test Set}} & 7.045 $\pm$ 1.653 & 7.008 $\pm$ 1.668 & 12.256 $\pm$ 5.424 & 36.226 $\pm$ 17.327\\ \hline
        \end{tabular}
        }
    \end{center}
    \caption{Average error and uncertainty measures on supershapes test sets.}\label{tab:SSresults}
\end{table}

\begin{figure}[!ht]
    \centering
    \includegraphics[width=\textwidth]{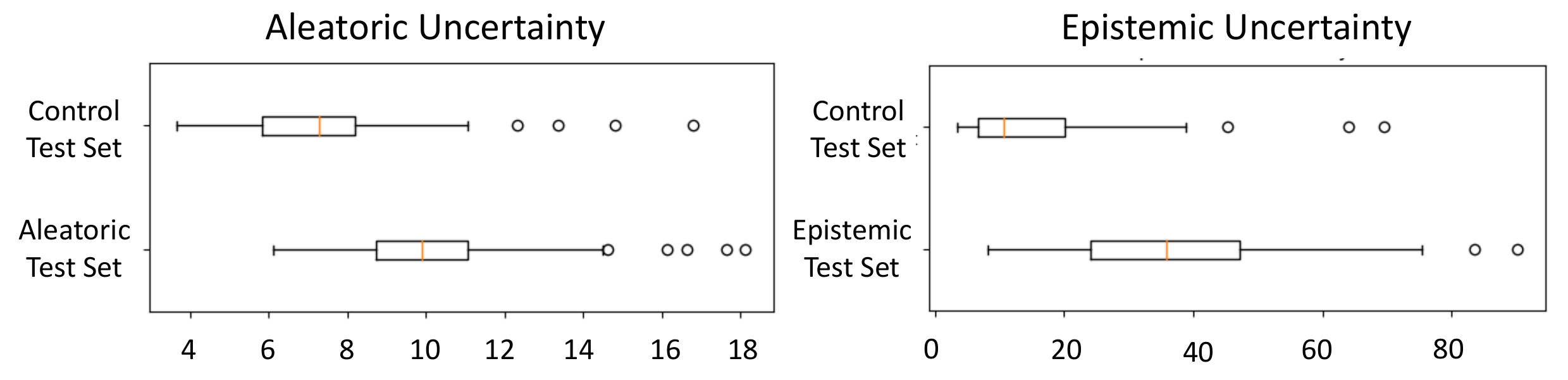}
    \caption{Boxplots of supershapes uncertainties compared to control test set.}\label{fig:SSboxplots}
\end{figure}

The results of all three test sets are shown in Table \ref{tab:SSresults}. The predictions of \UncertainDeepSSM are more accurate than DeepSSM on all of the test sets, with the aleatoric set having a notable difference. This is a result of the averaging effect of prediction sampling, which counters the effect of image blurring. The box plots of the uncertainty measure associated with the predicted PCA score in Fig. \ref{fig:SSboxplots} demonstrate that as expected, \UncertainDeepSSM predicts higher aleatoric uncertainty on the aleatoric test set and higher epistemic uncertainty on the epistemic test set when compared to the control. The epistemic test has the highest of both forms of uncertainty because changing the shape group produces a great shift in the image domain (aleatoric) and shape domain (epistemic).

\subsection{Left Atrium (LA) Dataset}

The LA dataset consists of 206 late gadolinium enhancement MRI images of size $235 \times 138 \times 175$ that vary significantly in intensity and quality and have surrounding anatomical structures with similar intensities.
The LA shape variation is also significant due to the topological variants pertaining to pulmonary veins arrangements \cite{ho2012left}.
The variation in images and shapes suggest a strong need for uncertainty measures.
For networks training purposes, the images are down-sampled to size $118 \times 69 \times 88$. We predict 19 PCA scores such that 95\% of the shape-population variability is preserved. 
\begin{figure}[!ht]
    \centering
    \includegraphics[width=\textwidth]{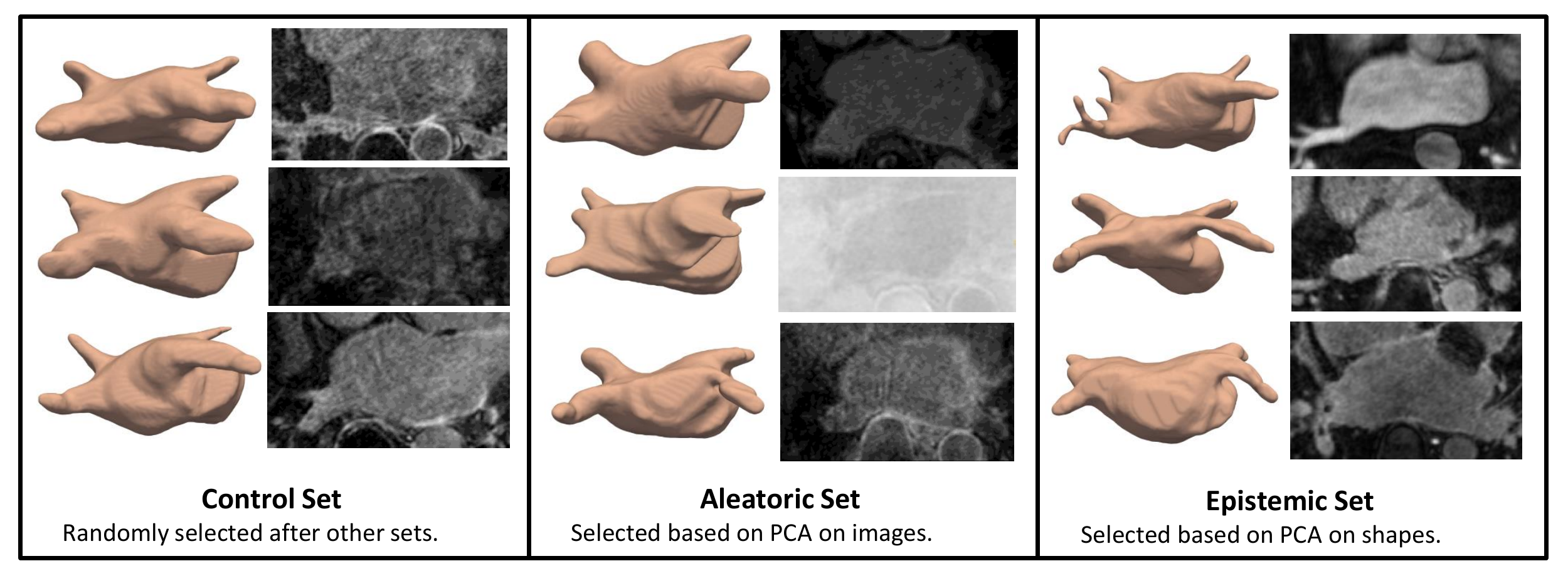}
    \caption{Examples from of surfaces and image slices for LA test sets.}
    \label{fig:LA}
\end{figure}
We compare DeepSSM and \UncertainDeepSSM on three test sets, each of size 30. To define the aleatoric test set, we run PCA (preserving 95\% of variability) on all 206 images. We then consider the Mahalanobis distance of the PCA scores of each sample to the mean PCA scores (within-subspace distance) as well as the image reconstruction error (mean square error as off-subspace distance). These values are normalized and summed to get a measure of image similarity to the whole set (similar to \cite{moghaddam1997probabilistic}). We select the 30 that differ the most to be the aleatoric test set. These examples are the least supported by the input data, suggesting they should have high data uncertainty. To define the epistemic test set, we use the same technique but perform PCA on the signed distance transforms, as an implicit form of shapes, rather than the raw images. In this way, we are able to select an epistemic test set of 30 examples with shapes that are poorly supported by the data. This selection technique produces aleatoric and epistemic test sets that overlap by 6 examples, leaving 152 out of 206 samples. 30 of these are randomly selected to be the control test set and the rest (122) are used in data augmentation to create a training set of 4000 and validation set of 1000. Examples from the test sets can be seen in Fig. \ref{fig:LA}. 
\begin{table}[]
\resizebox{\textwidth}{!}{%
\begin{tabular}{cc|c|c|c|c|}
\cline{3-6}
\multicolumn{1}{l}{}  &  & \multicolumn{1}{c|}{\textbf{DeepSSM}} & \multicolumn{3}{c|}{\textbf{\UncertainDeepSSM}} \\ \cline{3-6} 
\multicolumn{1}{l}{}  &  & \multicolumn{1}{c|}{\begin{tabular}[c]{@{}c@{}}\textbf{Surface-to-Surface }\\ \textbf{Distance} (mm) \end{tabular}} & \multicolumn{1}{c|}{\begin{tabular}[c]{@{}c@{}}\textbf{Surface-to-Surface} \\ \textbf{Distance} (mm) \end{tabular}} & \multicolumn{1}{c|}{\begin{tabular}[c]{@{}c@{}}\textbf{Aleatoric} \\ \textbf{Uncertainty}\end{tabular}} & \multicolumn{1}{c|}{\begin{tabular}[c]{@{}c@{}}\textbf{Epistemic}\\  \textbf{Uncertainty}\end{tabular}} \\ \hline
\multicolumn{1}{|c|}{\textbf{Control}}    & 25\% Train & 15.262 $\pm$ 3.694 & 10.670 $\pm$ 2.560 & 519.026 $\pm $7.357 & 58.206 $\pm$ 38.145\\
\multicolumn{1}{|c|}{\textbf{Test Set}}  & 75\% Train & 10.319 $\pm$ 2.834 & 10.072 $\pm$ 2.812 & 452.025 $\pm$ 0.519 & 46.581 $\pm$ 32.678 \\
\multicolumn{1}{|l|}{}                   & 100\% Train& 10.205 $\pm$ 2.779 & 10.153 $\pm$ 2.904& 431.518 $\pm$ 0.674 & 43.561 $\pm$ 29.821 \\\hline
\multicolumn{1}{|c|}{\textbf{Aleatoric}} & 25\% Train & 12.967 $\pm$ 3.592 & 12.830 $\pm$ 3.543 & 472.359 $\pm$ 10.164 & 64.312 $\pm$ 45.656\\
\multicolumn{1}{|c|}{\textbf{Test Set}}  & 75\% Train & 12.507 $\pm$ 3.522 & 12.169 $\pm$ 3.493 & 465.951 $\pm$ 1.089 & 60.458 $\pm$ 45.454\\
\multicolumn{1}{|l|}{}                   & 100\% Train& 12.242 $\pm$ 3.602 & 12.289 $\pm$ 3.525 & 442.129 $\pm$ 0.917 & 56.009 $\pm$ 42.371\\ \hline
\multicolumn{1}{|c|}{\textbf{Epistemic}} & 25\% Train & 15.759 $\pm$ 4.301 & 14.975 $\pm$ 4.209 & 465.817 $\pm$ 7.567 & 75.854 $\pm$ 51.188\\
\multicolumn{1}{|c|}{\textbf{Test Set}}  & 75\% Train & 14.690 $\pm$ 4.166 & 14.581 $\pm$ 4.104 & 446.641 $\pm$ 1.127 & 64.236 $\pm$ 44.642 \\
\multicolumn{1}{|l|}{}                   & 100\% Train& 14.558 $\pm$ 4.151 & 14.465 $\pm$ 4.092 & 448.082 $\pm$ 1.291 & 61.517 $\pm$ 42.259 \\ \hline
\end{tabular}
}
\caption{Results on left atrium test sets with various training set sizes. Reported surface-to-surface distances are averaged across the test set and uncertainty measures are averaged across PCA modes and the test set.}
\label{tab:LAresults}
\end{table}

\begin{figure}[!ht]
    \centering
    \includegraphics[width=\textwidth]{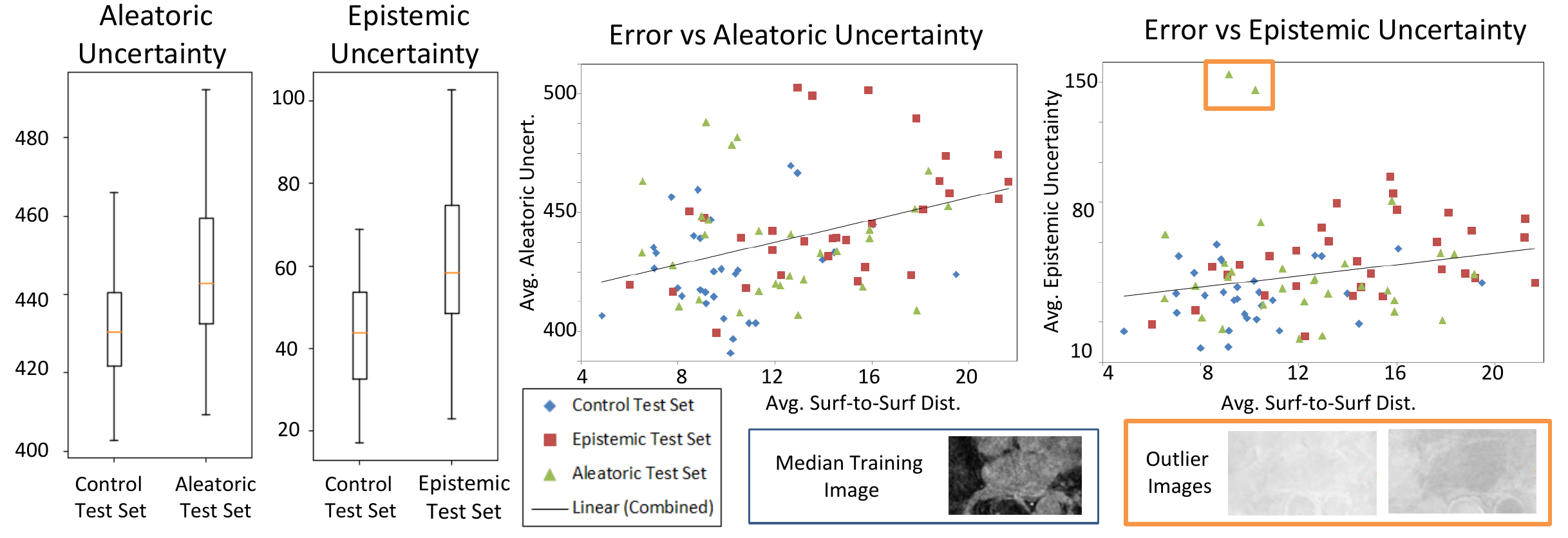}
    \caption{Results on LA test sets from training \UncertainDeepSSM on the full training set. The box plots show output uncertainty measures compared to the control test set. The scatter plot shows the average error versus average uncertainty on all three test sets. The two outliers marked with an orange box in the epistemic uncertainty plot are examples with images of a much higher intensity than the training examples (shown to the right) causing a spike in epistemic uncertainty. }
    \label{fig:LAplots}
\end{figure}

We train both DeepSSM and \UncertainDeepSSM on different percentages of training data, namely 100\%, 75\%, and 25\%, where an X\% is randomly drawn from the remaining 122 samples and then used to proportionally augment the data.
The average results of these tests are shown in Table \ref{tab:LAresults}. As expected, epistemic uncertainty measures decrease with more training data because model uncertainty can be explained away given more data. \UncertainDeepSSM made more accurate predictions in most cases, notably when training data is limited. \UncertainDeepSSM also successfully quantified uncertainty as we can see in the box plots in Fig. \ref{fig:LAplots}, which illustrate increased uncertainty measures on the uncertain test sets as compared to the control. The scatter plot in Fig. \ref{fig:LAplots} illustrates the correlation between accuracy and uncertainty measures. The trend lines (combined based on all three test sets) indicate that the uncertainty quantification from \UncertainDeepSSM provides insight into how trustworthy the model output is.
\newpage

In Fig. \ref{fig:teaser}, the uncertainty measures are shown on the meshes constructed from model predictions (models trained on 100\% of the training data). The top example is from the control set, the second is from the aleatoric set, and the bottom is from the epistemic set. Here, we can see that both aleatoric and epistemic uncertainty are higher in regions where the surface-to-surface distance is higher. This demonstrates the practicality of \UncertainDeepSSM in a clinical setting as it indicates what regions of the predicted shape professionals can trust and where they should be skeptical.
\section{Conclusion}

\UncertainDeepSSM provides a unified framework to predict shape descriptors with measures of both forms of uncertainty directly from 3D images. 
It maintains the end-to-end nature of DeepSSM while providing an accuracy improvement and uncertainty quantification.
By predicting and quantifying uncertainty on PCA scores, \UncertainDeepSSM enables population-level statistical analysis with aleatoric and epistemic uncertainty measures that can be evaluated in a visually interpretable way. 
In the future, a layer that maps the PCA scores to the set of correspondence points could be added, enabling fine-tuning the network and potentially providing an accuracy improvement over deterministically mapping PCA scores. 
\UncertainDeepSSM bypasses the time-intensive and cost-prohibitive steps of traditional SSM while providing the safety measures necessary to use deep network predictions in clinical settings. Thus, this advancement has the potential to improve medical standards and increase patient accessibility.

\newpage
\section{Acknowledgement}

This work was supported by the National Institutes of Health under grant numbers NIBIB-U24EB029011, NIAMS-R01AR076120, NHLBI-R01HL135568, and NIGMS-P41GM103545. The content is solely the responsibility of the authors and does not necessarily represent the official views of the National Institutes of Health.
MRI scans and segmentation were obtained retrospectively from the AFib database at the University of Utah.
The authors would like to thank the Division of Cardiovascular Medicine (data were collected under Nassir Marrouche, MD, oversight and currently managed by Brent Wilson, MD, PhD) at the University of Utah for providing the left atrium MRI scans and their corresponding segmentations.

\bibliographystyle{splncs04}
\bibliography{references}
\end{document}